\documentclass[times, review, 10pt]{elsarticle}
\usepackage[numbers]{natbib}
\usepackage{float}
\usepackage{graphicx}
\usepackage{subfigure}
\usepackage{diagbox} % Package for diagonal line in table cells
\usepackage{soul}

\usepackage{xcolor}
\usepackage{comment}
\usepackage{booktabs}
\usepackage{graphicx} % Optional, for resizing

\definecolor{teal}{rgb}{0.0, 0.5, 0.5}
\definecolor{fdblue}{RGB}{74,111,197}
\definecolor{ldorange}{RGB}{242,120,40}

\usepackage{amsmath,amssymb,amsfonts}

\usepackage{hyperref}

\hypersetup{
    colorlinks=true,
    linkcolor=blue,
    filecolor=blue,      
    urlcolor=blue,
    citecolor=blue,
}

\usepackage{caption}
\newcommand\blfootnote[1]{%
  \begingroup
  \renewcommand\thefootnote{}\footnote{#1}%
  \addtocounter{footnote}{-1}%
  \endgroup
}

\def\,{$\mskip\thinmuskip$} \def\!{$\mskip-\thinmuskip$}

\usepackage{algorithm}
\usepackage{algpseudocode}
\usepackage{amsmath}
\usepackage{multirow}
\usepackage{array} % for better alignment control
\def\BibTeX{{\rm B\kern-.05em{\sc i\kern-.025em b}\kern-.08em
    T\kern-.1667em\lower.7ex\hbox{E}\kern-.125emX}}

\begin{document}
\title{Dependency-aware synthetic tabular data generation}

\author[1]{Chaithra Umesh}
\author[1]{Kristian Schultz}
\author[1]{Manjunath Mahendra}
\author[1,4]{Saptarshi Bej}
\author[1,2,3]{Olaf Wolkenhauer}

\affiliation[1]{Institute of Computer Science, University of Rostock, Germany}
\affiliation[2]{Leibniz-Institute for Food Systems Biology, Technical University of Munich, Freising, Germany}
\affiliation[3]{Stellenbosch Institute for Advanced Study, South Africa}
\affiliation[4]{School of Data Science, Indian Institute of Science Education and Research, Thiruvananthapuram, India}

\begin{abstract}
Synthetic tabular data is increasingly used in privacy-sensitive domains such as healthcare, but existing generative models often fail to preserve inter-attribute relationships. In particular, functional dependencies (FDs) and logical dependencies (LDs), which capture deterministic and rule-based associations between features, are rarely or often poorly retained in synthetic datasets. To address this research gap, we propose the Hierarchical Feature Generation Framework (HFGF) for synthetic tabular data generation. We created benchmark datasets with known dependencies to evaluate our proposed HFGF. The framework first generates independent features using any standard generative model, and then reconstructs dependent features based on predefined FD and LD rules. Our experiments on four benchmark datasets with varying sizes, feature imbalance, and dependency complexity demonstrate that HFGF improves the preservation of FDs and LDs across six generative models, including \textit{CTGAN}, \textit{TVAE}, and \textit{GReaT}. Our findings demonstrate that HFGF can significantly enhance the structural fidelity and downstream utility of synthetic tabular data. 
\blfootnote{Code is available at \url{https://github.com/Chaithra-U/HFGF}}
\end{abstract}

\maketitle


\section{Introduction}\label{sec:introduction}

\textbf{Inter-attribute dependencies in tabular data} are structured relationships that exist between features, extending beyond simple statistical correlations or associations. These relationships are defined at the row level and are essential for preserving the semantic and structural integrity of the dataset. One prominent type of relationship that has been extensively studied is \textbf{Functional Dependency (FD)}, which is commonly utilized in database normalization to break down large tables into smaller, more organized ones \cite{lee_information-theoretic_1987, liu_discover_2012}. 
An FD exists when the value of one or more attributes uniquely determines the value of another attribute \cite{xu_are_2024, umesh_preserving_2025}. These dependencies can be categorized based on their nature and direction, leading to classifications such as one-to-one and many-to-one. In addition to FDs, the concept of \textbf{Logical Dependency (LD)} has also been introduced, which encompasses rule-based constraints between features that are not strictly deterministic but generally hold within specific domains \cite{umesh_preserving_2025, long_llm-tabflow_2025}. For example, in clinical datasets, there is a logical dependency between sex and pregnancy status, as biologically male individuals cannot be pregnant \cite{umesh_preserving_2025}.

Researchers have developed various tools to identify and analyze FDs in tabular data \cite{buranosky_fdtool_2019,papenbrock_hybrid_2016, yao_mining_2008,wei_towards_2023}. Our previous research introduced a novel $Q$-function to measure inter-attribute LDs \cite{umesh_preserving_2025}. However, despite these advancements, researchers have not thoroughly investigated the explicit modeling and preservation of FD and LD during synthetic data generation. A gap remains in the research, emphasizing the need for approaches that preserve these dependencies when creating synthetic datasets.

\textbf{Research gap:} Existing generative models struggle to preserve both FDs and LDs. Recent advancements in synthetic data generation have made significant strides, yet a key challenge remains preserving the intricate dependencies between various attributes \cite{long_llm-tabflow_2025}. In prior research \cite{umesh_preserving_2025}, we conducted a comparative analysis of seven generative models to evaluate their effectiveness in maintaining FDs and LDs within synthetic datasets. To assess these dependencies in both real and synthetic data, we utilized tools such as \textit{FDTool} \cite{buranosky_fdtool_2019} and the $Q$-function \cite{umesh_preserving_2025}. This analysis, conducted on five publicly available datasets, revealed that while certain models preserved logical dependencies, none were able to simultaneously maintain both logical and functional dependencies. Additionally, a recent review \cite{r_navigating_2024} highlighted a gap in the ability of existing generative models to maintain intricate relationships, underscoring the need for more sophisticated approaches to tackle this limitation. This study introduces a framework for generating synthetic tabular data while preserving FDs and LDs.

\begin{figure}[ht]
    \centering
    \includegraphics[width=\textwidth]{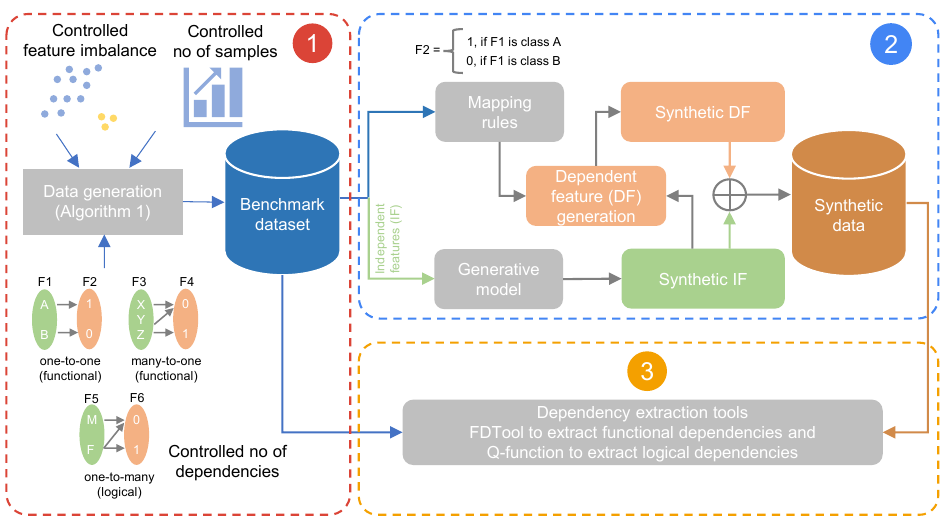}
    \caption{Hierarchical feature generation framework to improve the preservation of FDs and LDs in synthetic data. The workflow includes: (1) generating benchmark datasets under controlled conditions, (2) using generative models to synthesize independent features and mapping dependent features with known dependencies, and (3) evaluating dependency preservation using \textit{FDTool} and $Q$-function on both benchmark and synthetic datasets.}
    \label{workflow}
\end{figure}

\textbf{Our contribution:} 
% Briefly discuss your framework
We introduce a Hierarchical Feature Generation Framework (HFGF) designed to produce dependency-aware synthetic data to address the gap in preserving both FDs and LDs in synthetic data. The framework consists of two key steps (Box 2 in Figure \ref{workflow}). The first step focuses on identifying independent features in the dataset and generating synthetic independent features using generative models. In the second step, dependent features are mapped according to known dependency rules, ensuring that these dependencies are incorporated into the dependent synthetic data generation process. These independent and dependent synthetic features are subsequently concatenated to construct the final synthetic dataset. Our evaluation of the generated synthetic data demonstrates a notable improvement in the preservation of dependencies across all generative models when this hierarchical approach is applied. We assess the effectiveness of our proposed approach with a particular focus on benchmark datasets. Using benchmark datasets allows for precise control over dataset characteristics and enables a thorough evaluation of the proposed method's performance. A more detailed explanation of the methodology and results is provided in Section \ref{HFGF}.

\section{Related research} 
The focus of the present study is on the preservation of inter-attribute dependencies in synthetic tabular data. Recent advancements in research have led to the development of several methods for creating realistic tabular datasets. Notable models such as Generative Adversarial Networks (GANs) \cite{goodfellow_generative_2014}, Variational Autoencoders (VAEs) \cite{sami_comparative_2019}, diffusion models \cite{sattarov_findiff_2023}, convex space generators \cite{mahendra_convex_2024}, and Large Language Models (LLMs) \cite{zhao_survey_2023} have demonstrated their ability to accurately replicate the characteristics of real data. Each of these models employs distinct strategies and offers specific benefits to address the unique challenges inherent in tabular data generation. These challenges include managing diverse data types, maintaining the relationships between different columns, and managing high-dimensional data spaces \cite{figueira_survey_2022}.  A deeper understanding of the functionalities and applications of the models we employed in the study is discussed below.

\textit{CTGAN}, introduced by Xu \textit{et al.}\@ in 2019, generates synthetic tabular data using Conditional Generative Adversarial Networks. It employs mode-specific normalization to model non-Gaussian and multimodal continuous distributions \cite{xu_modeling_2019}. A variational Gaussian mixture model (VGM) identifies the number of modes for a continuous column and calculates the probability of each value belonging to a specific mode. A one-hot encoded mode and a normalized scalar represent continuous values. Discrete data uses one-hot encoding \cite{schultz_convgen_2022}. To address categorical feature imbalances, a training-by-sampling strategy ensures even sampling from all categories \cite{xu_modeling_2019}. A conditional vector selects a specific value for each discrete column based on the sum of cardinalities of the discrete columns. The Probability Mass Function (PMF) is calculated using the frequency of values in the column, and the conditional vector is adjusted accordingly \cite{schultz_convgen_2022}. The generator uses this conditional vector alongside a noise vector of random values. To prevent mode collapse, the concept of ‘packing’ allows the discriminator to process multiple samples simultaneously \cite{lin_pacgan_2018}. \textit{CTGAN} has been benchmarked with various datasets, demonstrating that it learns more accurate distributions than Bayesian networks \cite{xu_modeling_2019}. \par

\textit{CTABGAN+} is an improved version of the \textit{CTABGAN} algorithm, aimed at improving the quality of synthetic data for machine learning and statistical similarity. It introduces a new feature encoder for variables with a single Gaussian distribution and utilizes a revised min-max transformation for normalizing these variables \cite{zhao_ctab-gan_2024}. The algorithm includes a mixed-type encoder to handle categorical and continuous variables and manage missing values effectively. It employs the Wasserstein distance with a gradient penalty loss for more stable GAN training \cite{weng_gan_2019}. Additionally, an auxiliary classifier or regressor is integrated to boost performance in classification and regression tasks \cite{zhao_ctab-gan_2024}. \textit{CTABGAN+} uses a newly designed conditional vector based on log probabilities to tackle mode collapse in imbalanced data. It also incorporates Differential Privacy (DP) through the DP-SGD algorithm, simplifying training using a single discriminator \cite{zhao_ctab-gan_2024}. \textit{CTABGAN+} demonstrates superior synthetic data quality, showing higher utility and similarity than ten baseline methods across seven tabular datasets \cite{zhao_ctab-gan_2024}.\par

\textit{TVAE} is a deep learning model that generates synthetic data using probabilistic techniques \cite{vahdat_nvae_2020}. It has an encoder and a decoder, which help manage the latent space and variability in the data. By employing a reparameterization strategy, VAEs maximize the likelihood of the observed data and minimize divergence between the latent distribution and a predefined prior, allowing for back-propagation through stochastic sampling \cite{kingma_introduction_2019}. While VAEs improve data augmentation, they struggle with discrete data and may face issues like information loss, posterior collapse, and sensitivity to prior distribution \cite{kingma_introduction_2019}.\par

\textit{NextConvGeN} generates synthetic tabular data through convex space learning, creating samples that resemble the original data while staying within its neighborhoods \cite{mahendra_convex_2024}. It leverages neighborhoods of closely located real data points to learn convex coefficients through an iterative process involving two neural networks. This tool utilizes a generator-discriminator architecture similar to \textit{ConvGeN} \cite{schultz_convgen_2022}, but is tailored for tabular datasets. The generator works with randomized neighborhoods of real data, determined by the Feature Distributed Clustering (FDC) method, which effectively organizes high-dimensional clinical data \cite{bej_accounting_2023}. The discriminator's role is to classify synthetic points against shuffled data outside the input neighborhood, enhancing classification performance.\par

\textit{TabuLa}, created by Zilong Zhao \textit{et al.}\@ in 2025, is an LLM-based framework designed to speed up tabular data synthesis. Unlike existing models like \textit{GReaT} \cite{borisov_language_2023} and REaLTabFormer \cite{solatorio_realtabformer_2023} that rely on extensive training with pre-trained models, \textit{TabuLa} introduces four key innovations: i) It uses a randomly initialized language model for quicker adaptation to tabular data \cite{zhao_tabula_2025}. ii) It optimizes a foundational model from scratch, specifically for tabular tasks \cite{zhao_tabula_2025}. iii) It reduces token sequence length by consolidating column names and categorical values into single tokens, which shortens training time and improves learning efficiency \cite{zhao_tabula_2025}. iv) It employs middle padding to maintain the absolute positions of features within data columns, enhancing the representation of tabular data \cite{zhao_tabula_2025}. Experimental results show that \textit{TabuLa} reduces training time per epoch by an average of $46.2$\% compared to other LLM-based algorithms and achieves better utility with synthetic data \cite{zhao_tabula_2025}. \par

\textit{GReaT} (Generation of Realistic Tabular Data) is a generative model based on a transformer-decoder architecture for heterogeneous tabular data \cite{borisov_language_2023}. It simplifies the process by converting each row into text, avoiding traditional preprocessing like encoding and scaling \cite{borisov_language_2023}. To handle the lack of inherent feature order in tabular data, \textit{GReaT} applies random feature permutations, maintaining order independence for flexible data generation. The model, fine-tuned from a pre-trained autoregressive language model, can generate new samples based on different conditioning levels \cite{borisov_language_2023}. These include generating samples using only feature names, conditioning on single feature values, or using multiple name-value pairs for more complex sampling \cite{borisov_language_2023}. \textit{GReaT} offers advantages such as probabilistic control over sampling, enhanced data representation through knowledge from large text databases, and no need for explicit preprocessing, making it a straightforward solution for synthetic tabular data generation \cite{borisov_language_2023}. \par

Recent advancements have led to the development of generative models like \textit{GOGGLE} (Generative Modeling with Graph LEarning) \cite{liu_goggle_2022}, \textit{TABSYN} \cite{zhang_mixed-type_2023}, \textit{LLM-TabFlow} \cite{long_llm-tabflow_2025}, \textit{TabDiff} \cite{shi_tabdiff_2024}, \textit{CuTS} (Customizable Tabular Synthetic Data Generation) \cite{vero_cuts_2024}, \textit{KAMINO} \cite{ge_kamino_2021}, and \textit{C3TGAN} \cite{han_c3-tgan-_2023}. These models are designed to preserve the relational structure of the data \cite{liu_goggle_2022} by learning the correlation between the features \cite{shi_tabdiff_2024, ge_kamino_2021}, inter-column relationships \cite{zhang_mixed-type_2023}, causal relationships \cite{zhang_causaldifftab_2025}, and logical constraints \cite{long_llm-tabflow_2025, vero_cuts_2024}. However, we did not include them in our comparison because they are mainly made for supervised tasks like classification and regression. This focus makes them less helpful in assessing unsupervised data without a specific target column. Some of these models have issues with their publicly available and reproducible code \cite{long_llm-tabflow_2025} and have been tested on larger datasets (more than $10000$ rows). Many generative models have emphasized utility, fidelity, and privacy, yet the preservation of inter-attribute dependencies remains under-explored \cite{r_navigating_2024}. Our previous comparative analysis of state-of-the-art models revealed that no model maintained both FDs and LDs in synthetic data \cite{umesh_preserving_2025}. To address this gap, we propose HFGF, designed to enhance the generation of synthetic tabular data while improving the preservation of FDs and LDs. We created benchmark datasets to test the framework in different conditions. Our investigation determines that using HFGF helps state-of-the-art generative models effectively maintain FDs and LDs. A more detailed explanation of this framework is provided in Section \ref{HFGF}.

\begin{algorithm}
\caption{Benchmark Data Generation with Dependencies}
\label{alg:generate_simulated_data}
\begin{algorithmic}[0]

\Statex \textbf{Definitions:}
\vspace{-1.0em}
\Statex \item $n$: number of rows in the benchmark dataset \\
\vspace{-2.0em}
\Statex \item $config$: dictionary describing metadata for each feature $f_i$, where $i = 1, \dots, m$  \\
\vspace{-2.0em}
\Statex \item $f_i.\text{type}$: feature type, one of \texttt{int}, \texttt{categorical}, \texttt{id}, \texttt{one\_to\_one}, \texttt{many\_to\_one}, \texttt{one\_to\_many} \\
\vspace{-2em}
\Statex \item $f_i.\text{min}$ and $f_i.\text{max}$: specify the numeric range \\
\vspace{-2em}
\Statex \item $f_i.\text{categories}$ and  $f_i.\text{probabilities}$: define categories and their distribution \\
\vspace{-2em}
\Statex \item $f_i.\text{dependent\_feature}$: feature on which $f_i$ is conditionally dependent ($f_j$) \\
\vspace{-2em}
\Statex \item $f_i.\text{mapping}$: specify how categories of $f_j$ maps to categories of $f_i$, with probabilities\\
\vspace{-2em}
\Statex \item $c$: categories of $f_j$ \\
 \vspace{-0.8em}
\Require 
    $n$ and $config$
\Ensure 
    $\text{Benchmark data} \; D$
\Procedure{GenerateBenchmarkData}{$n$, $config$}
    \State initialize random seed and empty dictionary $data \gets \{\}$
    
    \For{$f_i$ in $config.\text{features}$}
        \If{$f_i.\text{type} = \text{int}$}
            \State $f_i \gets \texttt{np.random.randint}(f_i.\text{min}, f_i.\text{max}, n)$
        \ElsIf{$f_i.\text{type} = \text{categorical}$}
            \State $f_i \gets \texttt{np.random.choice}(f_i.\text{categories}, n, p=f_i.\text{probabilities})$
        \ElsIf{$f_i.\text{type} = \text{id}$}
            \State $f_i \gets [\text{format}(i) \text{ for } i \in \{1, \dots, n\}]$
        \ElsIf{$f_i.\text{type} \in \{\text{one\_to\_one}, \text{many\_to\_one}\}$}
            \State retrieve $f_j \gets f_i. \text{dependent\_feature}$
            \State $f_i \gets [f_i.\text{mapping}[category] \text{ for } category\in f_j]$
        \ElsIf{$f_i.\text{type} = \text{one\_to\_many}$}
            \State retrieve $f_j \gets f_i. \text{dependent\_feature}$
            \State $f_i \gets [\texttt{np.random.choice}(f_i.\text{mapping}[c].\text{categories},$ 
            \Statex \hspace{6.5em} $p=f_i.\text{mapping}[c].\text{probabilities}) \text{ for } c \in f_j]$
        \EndIf
        \State store $f_i$ in $data$
    \EndFor
    \State \Return $\text{Benchmark data} \; D$
\EndProcedure
\end{algorithmic}
\end{algorithm}
 
\section{Hierarchical Feature Generation Framework} \label{HFGF}
HFGF consists of three key steps, as outlined in the Figure \ref{workflow}. The first step involves the generation of benchmark data, detailed in the Algorithm \ref{alg:generate_simulated_data}. To begin, the user has to specify the desired number of rows, $n$, and define a configuration dictionary $config$. This dictionary encodes metadata for each feature in the dataset. For numerical features, the metadata specifies the minimum and maximum allowable values. For categorical features, it comprises a set of categories and their corresponding distributions. The $config$ dictionary also outlines inter-attribute dependencies, specifying which features are dependent on others and the nature of these dependencies. These dependencies are categorized as either functional (one-to-one and many-to-one) or logical (one-to-many), with explicit mapping rules used during data generation to enforce them. By adjusting the $config$ dictionary and the value of $n$, HFGF facilitates the creation of diverse benchmark datasets with controlled structural properties.

\subsection{Experimental protocols}
There are no well-established benchmark datasets that include both FDs and LDs to evaluate our proposed framework. To address this, we generated four benchmark datasets under various conditions for our analysis.
\begin{itemize}
    \item A dataset with $100$ rows, consisting of $7$ features, $5$ categorical, $1$ integer, and $1$ identifier with $4$ FDs and $64$ LDs
    \item A dataset with $1000$ rows and the same feature and dependency structure as in 1 
    \item A dataset with $100$ rows and $8$ features, $6$ categorical, $1$ integer, and $1$ identifier, exhibiting $7$ FDs and $84$ LDs
    \item A complex dataset with $100$ rows and $15$ features, $13$ categorical and $1$ numerical, $1$ identifier with $40$ FDs and $324$ LDs, additionally incorporating imbalance in two categorical features
\end{itemize}

In the second step of HFGF, synthetic data is generated using six generative models: \textit{CTGAN} and \textit{CTABGAN+} (GAN-based), \textit{TVAE} (variational autoencoder-based), \textit{NextConvGeN} (convex space-based), \textit{TabuLa}, and \textit{GReaT} (transformer-based). Rather than modeling all features, we restrict the generative process to only the independent features, i.e., those that are not dependent on any other feature in the configuration dictionary. We define a feature to be independent if it does not appear as the right-hand side (RHS) of any dependency and also does not act as a dependent in any other relationship. This is critical in datasets with multiple overlapping dependencies, where some features may serve as both LHS and RHS in different relationships. Such features are excluded from the independent set. In case of one-to-one FDs (i.e., bijective mappings), where two features deterministically imply each other, either feature may be treated as independent, since generating one enables deterministic reconstruction of the other. Once independent features are identified (for benchmark data, we already know which features are dependent and which are independent; for real data, extract dependencies using \textit{FDTool} and identify dependent and independent features), synthetic independent features are generated using the chosen generative model. Dependent features are reconstructed by applying the mapping rules specified in the configuration dictionary. These mappings are specified explicitly for each dependent feature as shown in the second step of Figure \ref{workflow}. This mapping indicates that \textsf{F2} is deterministically derived from \textsf{F1} based on the specified correspondence. The reconstruction process is applied recursively to all dependent features, preserving the defined FDs and LDs. For benchmark datasets, dependency mappings are predefined in the configuration file. For real-world datasets, we infer these mappings using \textit{FDTool} and identify dependent and independent features accordingly. The final synthetic dataset is obtained by concatenating the synthetic independent features with the deterministically reconstructed dependent features.\par

The third step evaluates the extent to which the generated synthetic data preserves FDs and LDs across the six generative models. We utilized \textit{FDTool} \cite{buranosky_fdtool_2019} to identify FDs and the $Q$-function \cite{umesh_preserving_2025} for LDs. $Q$-function is defined as:
\begin{equation}
\begin{array}{lll}
 Q_T(\mathcal{A}, \mathcal{B} ) & = \frac{|\{ (a,b) \colon a \in A, b \in B \text{ and } a \sim_T b \}| - |A|}{|A| \cdot (|B| - 1)} &  \text{if } |A| \geq 1 \text{ and }|B| > 1 \\
 Q_T(\mathcal{A}, \mathcal{B})   & = 0  &  \text{if } |A| = 0 \text{ or } |B| \leq 1 
\end{array}
\label{q-function}
\end{equation}
where $\mathcal{A}$ and $\mathcal{B}$ are subsets of columns (attributes or features) selected from the total columns $C$ of the dataset $T$. $A$ and $B$ are the sets of all unique tuples that exist in the table $T$ for the given column selections $\mathcal{A}$ and $\mathcal{B}$. $a \in A$ refers to an individual unique tuple from the set $A$.  $b \in B$ refers to an individual unique tuple from the set $B$. $a \sim_T b$ is a relation that holds if and only if $a \in A$ and $b \in B$ are in the same row in the table $T$. $C^* \subseteq C$ is the set containing all possible selections of columns. The $Q$-function $Q_T: C^* \times C^* \to [0,1]$ provides $Q$-scores between $0$ and $1$ for every pair of column selections $\mathcal{A} \in C^*$ and $\mathcal{B} \in C^*$ within the dataset. A score of $0$ in the $Q$ function indicates that the attributes are functionally dependent on each other. If the score is $1$, then the attributes are not dependent on each other. The attributes are logically dependent if the score is between $0$ and $1$.  We refer to \cite{umesh_preserving_2025} for a detailed mathematical explanation of the $Q$-function. Overall, the HFGF framework significantly improves the preservation of dependencies in synthetic data. The algorithm to generate  benchmark datasets is available at \url{https://github.com/Chaithra-U/HFGF}.

% make sure to end this paragraph with algorith availability instead this
%Our analysis indicated that GAN-based models and the \textit{GReaT} model demonstrated enhanced performance in preserving inter-attribute dependencies across all scenarios when utilizing HFGF (See Table \ref{case1&2} and \ref{case3&4}). Conversely, while \textit{NextConvGeN} and \textit{TabuLa} maintained dependencies effectively without the framework in the initial three scenarios, their performance declined under increased dependency and feature imbalance without HFGF (Refer to Table \ref{next&tab}). Overall, our framework significantly improves the preservation of dependencies in synthetic data.

\section{Results}\label{sec: results} 

The algorithm to generate benchmark datasets can be used to test synthetic tabular data generative models with respect to their ability to preserve inter-attribute functional and logical dependencies (FD and LD) in tabular data. We here used four benchmark datasets to test and compare six state-of-the-art generative models for synthetic tabular data generation. The validation study demonstrates the value of the proposed Hierarchical Feature Generation Framework (HFGF).\par

\begin{figure}[ht]
    \centering
    \includegraphics[width=\textwidth]{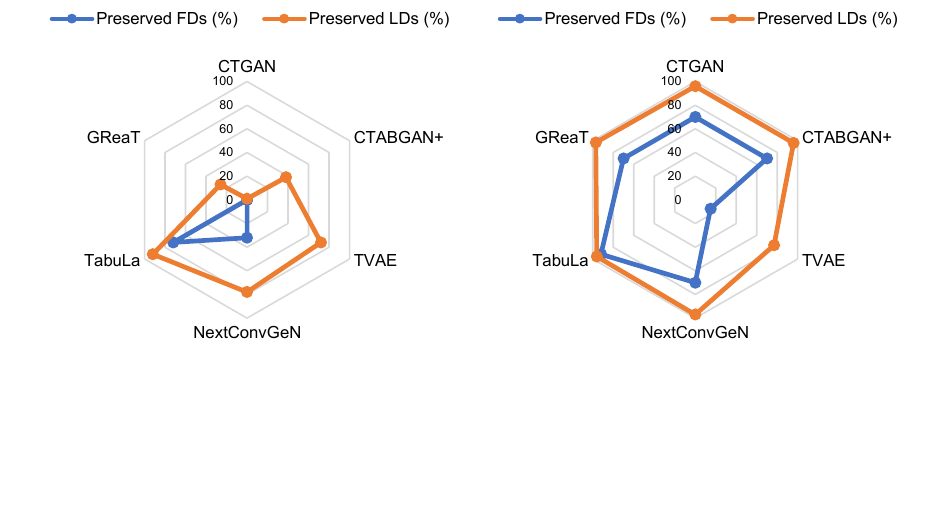}
    \vspace{-7.50em}
    \caption{Radar chart comparing the preservation of dependencies with and without Hierarchical Feature Generation Framework (HFGF) using six generative models. Preservation is computed as the percentage of benchmark data FDs/LDs that are also discovered in the synthetic data using \textit{FDTool} and the $Q$-function. The evaluation is based on the fourth benchmark dataset, which includes $40$ FDs, $324$ LDs, and exhibits feature imbalance. Incorporation of HFGF consistently improves dependency preservation across all models.}
    \label{spider}
\end{figure}

\textbf{HFGF improves preservation of FDs and LDs in synthetic tabular data:}
Generating tabular data using state-of-the-art generative models fails to preserve both FDs and LDs. Incorporating HFGF with generative models improves both FDs and LDS in synthetic tabular data. Figure \ref{spider} compares the preservation of percentages of FDs (blue line) and LDs (orange line) across six generative models: \textit{CTGAN}, \textit{CTABGAN+}, \textit{TVAE}, \textit{NextConvGeN}, \textit{TabuLa}, and \textit{GReaT}, without and with HFGF implementation. The radar chart on the left indicates that, except for the \textit{NextConvGeN} and \textit{TabuLa} models, all other generative models failed entirely to preserve any FDs. Conversely, with HFGF applied (right-side chart), all models demonstrate significant improvement in preserving FDs.\par

This discrepancy arises because generative models aim to generate synthetic data that resembles real data properties without compromising privacy, they were not designed to preserve FDs. However, preserving FDs in synthetic data can be challenging, as even a single contradiction in a data point in the synthetic data disregards two attributes in the synthetic dataset as an FD. In contrast, HFGF improves the situation by focusing on mapping dependent features rather than merely generating them, thereby significantly increasing the likelihood of retaining both FDs and LDs. While some models can preserve certain LDs in the absence of HFGF, the integration of this framework enables them to maintain all LDs present in the real datasets. This study underscores the effectiveness of incorporating HFGF with generative models to enhance the preservation of inter-attribute dependencies, rather than relying solely on conventional generation approaches.

\textbf{GAN-based, \textit{TVAE}, and \textit{GReaT} models show consistent improvement across all four scenarios with HFGF:}
Tables \ref{case1&2} and \ref{case3&4} summarize the effects of HFGF across four different cases. In Case 1, none of the generative models effectively preserved FDs without applying HFGF. This limitation likely results from the small dataset size ($100$ rows), which challenges data-hungry models such as GANs (\textit{CTGAN}, \textit{CTABGAN+}) and transformers (\textit{GReaT}). However, incorporating HFGF significantly improved FD preservation for these models.
In Case 2, the dataset size increased to $1000$ rows, yet GAN-based and transformer-based models did not show notable improvements in FD preservation without HFGF. However, LD preservation slightly improved. With HFGF, these models preserved approximately $75$\% of FDs and all LDs (refer Table\ref{case1&2}, Case 2).\par

\begin{table}[ht]
\centering
\renewcommand{\arraystretch}{1.3} % optional: increase row height
\begin{tabular}{>{\centering\arraybackslash}m{2.3cm} | c c | c c || c c | c c}
\hline
\multirow{3}{=}{\textbf{Generative models}} & \multicolumn{4}{c||}{\textbf{Case 1}} & \multicolumn{4}{c}{\textbf{Case 2}} \\
\cline{2-9}
 & \multicolumn{2}{c|}{without HFGF} & \multicolumn{2}{c||}{with HFGF} & \multicolumn{2}{c|}{without HFGF} & \multicolumn{2}{c}{with HFGF} \\
\cline{2-9}
 & FDs & LDs & FDs & LDs & FDs & LDs & FDs & LDs \\
\hline
CTGAN     & 0  & 0   & \textcolor{fdblue}{75} & \textcolor{ldorange}{100} & 0  & 0   & \textcolor{fdblue}{75} & \textcolor{ldorange}{100} \\
CTABGAN+  & 0  & 31  & \textcolor{fdblue}{75} & \textcolor{ldorange}{100} & 0  & 0  & \textcolor{fdblue}{75} & \textcolor{ldorange}{100} \\
TVAE      & 0  & \textcolor{ldorange}{60}  & \textcolor{fdblue}{50} & 39  & 0  & 44  & \textcolor{fdblue}{75} & \textcolor{ldorange}{100}  \\
GReaT     & 0  & 9   & \textcolor{fdblue}{75} & \textcolor{ldorange}{100} & 0  & 81   & \textcolor{fdblue}{75} & \textcolor{ldorange}{100} \\
\hline
\end{tabular}
%\end{center}
\caption{Comparison of preserved FDs and LDs in Case 1 and Case 2, with and without HFGF. The results indicate that incorporating HFGF improves the preservation of both FDs and LDs across all models in both cases}
\label{case1&2}
\end{table}

In Case 3, additional categorical features are included. Even under these conditions, models using HFGF consistently outperformed those without, effectively preserving both FDs and LDs. In Case 4, the introduction of feature imbalance tested HFGF’s robustness using a small dataset ($100$ rows). Table \ref{case3&4}, Case 4 shows that all models incorporating HFGF preserved most of the FDs and LDs present in real data. \par
Notably, the \textit{TVAE} model consistently improved FD preservation across all scenarios when using HFGF. However, LD preservation is higher in Cases 1 and 3 without HFGF due to mode collapse, where \textit{TVAE} generates identical values for some features. This suggests that in some scenarios, HFGF’s reliance on accurate generation of independent features may amplify mode collapse effects, especially if those features are categorical and imbalanced. In HFGF, dependent features rely on independent features, and identical values in independent features can compromise logical dependency rules. Increasing the dataset size, particularly evident in Case 2, improved LD preservation with HFGF. Additionally, increased feature complexity in Case 4 further enhanced LD preservation. These results demonstrate that integrating HFGF consistently strengthens the preservation of inter-attribute dependencies in synthetic datasets, irrespective of dataset size, feature complexity, or feature imbalance.\par

\begin{table}[ht]
\centering
\renewcommand{\arraystretch}{1.3} % optional: increase row height
\begin{tabular}{>{\centering\arraybackslash}m{2.3cm} | c c | c c || c c | c c}
\hline
\multirow{3}{=}{\textbf{Generative models}} & \multicolumn{4}{c||}{\textbf{Case 3}} & \multicolumn{4}{c}{\textbf{Case 4}} \\
\cline{2-9}
 & \multicolumn{2}{c|}{without HFGF} & \multicolumn{2}{c||}{with HFGF} & \multicolumn{2}{c|}{without HFGF} & \multicolumn{2}{c}{with HFGF} \\
\cline{2-9}
 & FDs & LDs & FDs & LDs & FDs & LDs & FDs & LDs \\
\hline
CTGAN     & 0  & 0   & \textcolor{fdblue}{86} & \textcolor{ldorange}{100} & 0  & 1   & \textcolor{fdblue}{70} & \textcolor{ldorange}{96} \\
CTABGAN+  & 0  & 5  & \textcolor{fdblue}{86} & \textcolor{ldorange}{100} & 0  & 38  & \textcolor{fdblue}{70} & \textcolor{ldorange}{96} \\
TVAE    & 29  & \textcolor{ldorange}{71}  & \textcolor{fdblue}{57} & 33  & 0  & 72  & \textcolor{fdblue}{15} & \textcolor{ldorange}{77}  \\
GReaT     & 0  & 43   & \textcolor{fdblue}{86} & \textcolor{ldorange}{100} & 0  & 26   & \textcolor{fdblue}{70} & \textcolor{ldorange}{97} \\
\hline
\end{tabular}
\caption{Comparison of preserved FDs and LDs in Case 3 and Case 4, with and without HFGF. The results indicate that incorporating HFGF improves the preservation of both FDs and LDs across all models in both cases.}
\label{case3&4}
\end{table}

\textbf{\textit{NextConvGeN} and \textit{TabuLa} models exhibit increased FDs with HFGF, particularly in the case of complex dependencies and feature imbalance:}
The generation of synthetic data using the \textit{NextConvGeN} and \textit{TabuLa} models effectively preserves the majority of FDs and all LDs in the first three cases, even without using an HFGF. These specific cases involve fewer features, ranging from $7$ to $8$, which results in fewer dependencies overall. Both models demonstrate their capability to accurately capture inter-attribute relationships in such scenarios, as illustrated in Table \ref{next&tab}. Moreover, including HFGF maintains the same percentage of preserved FDs and LDs for these first three cases. However, it is notable that in Case 1 and 3, the number of preserved FDs for \textit{NextConvGeN} slightly decreases with the use of HFGF. This reduction is due to \textit{NextConvGeN}'s failure to generate unique patient IDs, resulting in not preserving the FD related to patient IDs, which ultimately impacts the overall percentage of FDs.\par

\begin{table}[ht]
\centering
\renewcommand{\arraystretch}{1.3} % optional: increase row height
\begin{tabular}{>{\centering\arraybackslash}m{2.3cm} | c c | c c || c c | c c}
\hline
\multirow{3}{=}{\textbf{}} & \multicolumn{4}{c||}{\textbf{NextConvGeN}} & \multicolumn{4}{c}{\textbf{TabuLa}} \\
\cline{2-9}
 & \multicolumn{2}{c|}{without HFGF} & \multicolumn{2}{c||}{with HFGF} & \multicolumn{2}{c|}{without HFGF} & \multicolumn{2}{c}{with HFGF} \\
\cline{2-9}
 & FDs & LDs & FDs & LDs & FDs & LDs & FDs & LDs \\
\hline
\textbf{Case 1}    & \textcolor{fdblue}{100}  & 100   & 75 & 100 & 100  & 100   & 100 & 100 \\
\textbf{Case 2}     & 75  & 100    & 75 & 100 & 75   & 100    & 75 & 100 \\
\textbf{Case 3}    & \textcolor{fdblue}{100}  & 100  & 86 & 100  & 100  & 100  & 100 & 100  \\
\textbf{Case 4}   & 33  & 78   & \textcolor{fdblue}{70} & \textcolor{ldorange}{97} & 73  & 93   & \textcolor{fdblue}{93} & \textcolor{ldorange}{96} \\
\hline
\end{tabular}
\caption{Comparison of preserved FDs and LDs across all cases with and without HFGF using \textit{NextConvGeN} and \textit{TabuLa}. Results indicate that \textit{NextConvGeN} and \textit{TabuLa} effectively preserve both FDs and LDs without HFGF in the first three cases, while in the fourth Case, preservation significantly improves with HFGF.}
\label{next&tab}
\end{table}

In contrast, Case 4 presents a more complex situation with $15$ features, which leads to an increase in dependencies and an imbalance in feature distribution. These challenges hinder the models' effectiveness in maintaining FDs and LDs without including HFGF. The findings indicate that when data involves a larger number of dependencies and exhibits feature imbalances, the use of HFGF during synthetic data generation improves the preservation of FDs and LDs.\par

\section{Discussion}
% 1) As number of samples increases \textit{TVAE} will not generate same value for all features 2) \textit{NextConvGeN} and TabuLa can preserve FDs and LDs for small data with less dependencies 3)If you have larger tabular data and the goal is to preserve dependencies in synthetic data then HFGF will improve the dependencies 4)also mention since you are working with clinical datasets and they are often small in size, I wanted to check HFGf on samll datasets

\textbf{HFGF effectiveness in small-data regimes:}   
Interestingly, GAN-based, VAE-based, and transformer-based models such as \textit{GReaT} demonstrate improved performance in preserving FDs and LDs when combined with HFGF, despite their typical reliance on large datasets. This behavior can be attributed to the design of the HFGF framework. Rather than generating the full joint distribution, HFGF separates the feature space into independent and dependent features. Only the independent features are synthesized using the generative model, the dependent features are subsequently reconstructed based on known functional and logical relationships.\par

\textbf{Dependency reconstruction mechanism:} In our experimental setting, the benchmark data is created with explicitly defined dependencies, enabling accurate mapping from independent to dependent features. As the number of independent features is small and these features are uncorrelated, the generative models are exposed to a lower-dimensional and structurally simpler distribution to learn. This simplifies the training process and reduces the model's complexity. Given that clinical datasets are often small, sparse, and privacy-sensitive, HFGF’s ability to ensure structural fidelity in such contexts is a key enabler for safe and reliable synthetic data usage in biomedical research. We therefore explicitly evaluated HFGF under small-data conditions, varying the number of features to assess its robustness. Furthermore, given that GANs and transformer-based models typically require extensive data to perform well, this experimental setup allowed us to test their ability to preserve inter-attribute dependencies when using HFGF.\par

\begin{figure}[ht]
    \centering
    \includegraphics[width=\textwidth]{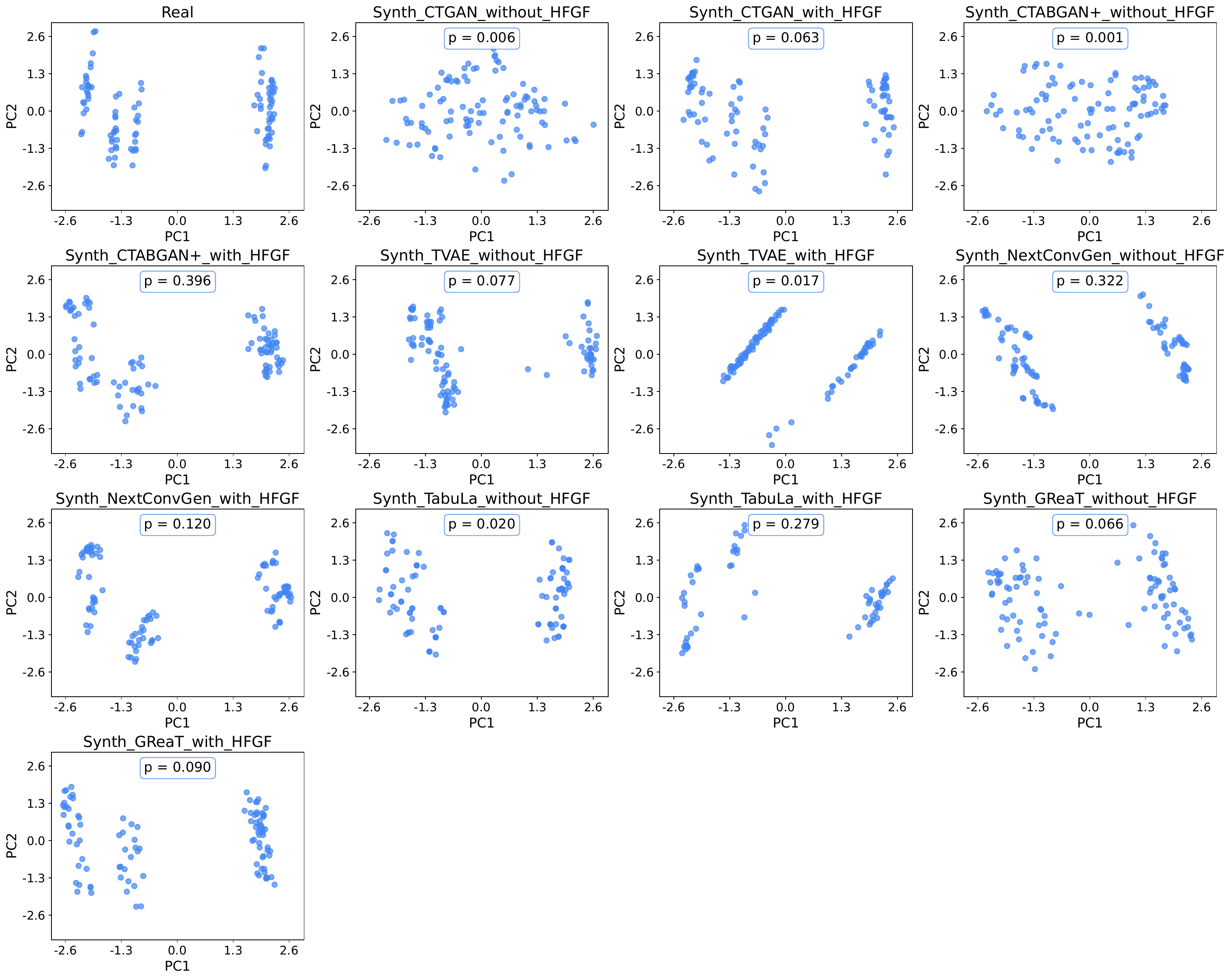}
    %\vspace{-7.50em}
    \caption{$2$-dimensional PCA embeddings of real data and synthetic data generated by six different generative models, with and without the use of HFGF, on fourth benchmark data. For each synthetic dataset, the corresponding $p$-value (shown above each plot) is computed using the Peacock test, which evaluates whether the real and synthetic embeddings are drawn from the same distribution. A $p$-value Greater than $0.05$ suggests no statistically significant difference between the distributions. Notably, models with HFGF not only improve dependency preservation but also align better with real data distributions, indicating that dependency preservation and distributional fidelity are not mutually exclusive.}
    \label{2d-case4}
\end{figure}

\textbf{Failure cases (e.g., missed class values in independent features):} Since both FDs and LDs are functions of the independent features, their preservation depends on the accurate generation of those independent variables. Provided that all categorical classes of an independent feature are represented in the generated data, the dependency structure can be reconstructed without loss. However, if the generative model fails to capture certain classes, especially in categorical independent features, then any dependent features relying on them may fail to preserve the intended dependency. In our study, only four independent features are generated, which proves to be tractable for the models even with limited training data. This partly explains the improved preservation performance observed with the HFGF framework. Quantitative results support this; combining HFGF with generative models leads to more preserved FDs and LDs than using the models alone.\par

\textbf{Limitations and future directions:} Although the proposed HFGF framework demonstrates improvements in preserving FDs and LDs, it exhibits certain limitations. Its applicability is dependent on clearly defined independent features in the dataset. In cases where all features are mutually dependent, the framework cannot reconstruct or preserve the underlying structure. Additionally, identifying pairwise dependencies is a quadratic process, and separating dependent from independent features adds a computational overhead that may affect scalability. Furthermore, the current simulation setup considers only dependencies among categorical features. Interactions between numerical and categorical features are not explicitly modeled. One possible extension involves discretizing numerical features into categorical bins.\par

Future work could focus on developing generative models that can explicitly learn and reproduce inter-attribute dependencies. One potential direction involves representing features and their dependencies as a graph, where nodes correspond to attributes and edges encode functional or logical relationships. This graph structure can then be incorporated into graph neural network (GNN) architectures, enabling synthetic tabular data to more comprehensively reflect the original dataset's statistical distributions and inter-attribute dependencies.

\textbf{Structural fidelity in lower-dimensional space:} We also evaluated whether the synthetic data preserved global distributional properties using principal component analysis (PCA). Figure \ref{2d-case4} demonstrates that the two-dimensional PCA projections of synthetic data generated with HFGF more closely resemble the real data distribution than those generated without HFGF. This observation is supported statistically via the Peacock test, a multivariate generalization of the Kolmogorov–Smirnov test, which compares the empirical cumulative distribution functions (ECDFs) of two multivariate samples. The null hypothesis is that the two samples originate from the same distribution. A $p$-value greater than $0.05$ indicates no statistically significant difference.\par

In most cases, synthetic data generated with HFGF has $p$-values above $0.05$, indicating statistical similarity to the real data distribution. Notably, for \textit{NextConvGeN}, $p$-values are higher in both with and without HFGF, suggesting that this model captures the underlying distribution well. However, the combination with HFGF still enhances the overall data dependencies (Table \ref{next&tab}) and structure preservation. Overall, these findings demonstrate that integrating HFGF with generative models facilitates the preservation of FDs and LDs and improves the alignment of synthetic and real data distributions in low dimensions.\par

\section{Conclusion}
Generating high-quality synthetic tabular data is important, especially in areas like clinical research, where balancing data utility and privacy is necessary. While advanced generative models do well at keeping overall data patterns and privacy intact, they struggle to maintain relationships between attributes, such as FDs and LDs. This study tackles this issue by introducing the HFGF, a new approach to better preserve FDs and LDs in synthetic datasets.\par

Our evaluation demonstrated that integrating HFGF with state-of-the-art generative models, \textit{CTGAN}, \textit{CTABGAN+}, \textit{TVAE}, and \textit{GReaT}, has consistently improved the preservation of both FDs and LDs across various benchmark datasets. This improvement is evident even with smaller datasets (e.g., $100$ rows), where traditional models often struggle. Beyond explicit dependency preservation, HFGF also maintained similar distributional characteristics in two-dimensional embeddings (PCA), addressing a gap in current methodologies that lack a dedicated focus on preserving inter-attribute dependencies. \par

The utility of synthetic data is diminished if the inter-attribute dependencies are not preserved for the downstream analysis, like patient stratification. HFGF mitigates this by providing a systematic mechanism to incorporate FDs and LDs into the generation process. It is essential to acknowledge that the current iteration of HFGF primarily acts as a framework to identify and map existing dependencies from real data onto synthetic data, rather than inherently learning these complex relationships within the generative model itself. In conclusion, for applications where preserving FDs and LDs is important, integrating HFGF into existing generative models offers a solution, enabling the reliability of synthetic data for downstream tasks.

\section*{CRediT authorship contribution statement}
\textbf{Chaithra Umesh:} Experimented and wrote the first version of the manuscript. \textbf{Kristian Schultz:} Discussed and reviewed the mathematical part of the manuscript. \textbf{Manjunath Mahendra:} Discussed and revised the manuscript's contents. \textbf{Saptarshi Bej:} Conceptualized, reviewed, and supervised the manuscript writing and experiments. \textbf{Olaf Wolkenhauer:} Reviewed and supervised the manuscript writing and experiments.

\section*{Conflict of Interest}
The authors have no conflict of interest.

\section*{Availability of code and results}
We provided detailed Jupyter notebooks from our experiments in \href{https://github.com/Chaithra-U/HFGF}{GitHub} to support transparency, re-usability, and reproducibility.

\section*{Acknowledgment}
This work has been supported by the German Research Foundation (DFG), FK 515800538, obtained for `Learning convex data spaces for generating synthetic clinical tabular data'.

\bibliographystyle{elsarticle-num}
%\bibliographystyle{unsrtnat}
%\bibliography{FunctionalDependencies}
\bibliography{main}

\end{document}